\documentclass{article}

\usepackage{microtype}
\usepackage{graphicx}
\usepackage{subcaption}
\usepackage{booktabs} 

\usepackage{hyperref}

\usepackage[preprint]{icml2026}

\usepackage{amsmath}
\usepackage{amssymb}
\usepackage{mathtools}
\usepackage{amsthm}

\usepackage[capitalize,noabbrev]{cleveref}

\theoremstyle{plain}

\theoremstyle{definition}

\theoremstyle{remark}

\usepackage[textsize=tiny]{todonotes}

\icmltitlerunning{FNF: Functional Network Fingerprint for Large Language Models}

\begin{document}

\twocolumn[
  \icmltitle{FNF: Functional Network Fingerprint for Large Language Models }



  \icmlsetsymbol{equal}{*}

  \begin{icmlauthorlist}
    \icmlauthor{Yiheng Liu}{1}
    \icmlauthor{Junhao Ning}{1}
    \icmlauthor{Sichen Xia}{1}
    \icmlauthor{Haiyang Sun}{1}
    \icmlauthor{Yang Yang}{1}
    \icmlauthor{Hanyang Chi}{1}
    \icmlauthor{Xiaohui Gao}{1}
    \icmlauthor{Ning Qiang}{2}
    \icmlauthor{Bao Ge}{2}
    \icmlauthor{Junwei Han}{1}
    \icmlauthor{Xintao Hu}{1}
  \end{icmlauthorlist}

  \icmlaffiliation{1}{School of Automation, Northwestern Polytechnical University, Xi'an, China}
  \icmlaffiliation{2}{School of Physics and Information Technology, Shaanxi Normal University, Xi'an, China}

  \icmlcorrespondingauthor{Xintao Hu}{xhu@nwpu.edu.cn}

  \icmlkeywords{Machine Learning, ICML}

  \vskip 0.3in
]



\printAffiliationsAndNotice{}  

\begin{abstract}
  The development of large language models (LLMs) is costly and has significant commercial value. Consequently, preventing unauthorized appropriation of open-source LLMs and protecting developers' intellectual property rights have become critical challenges. In this work, we propose the Functional Network Fingerprint (FNF), a training-free, sample-efficient method for detecting whether a suspect LLM is derived from a victim model, based on the consistency between their functional network activity. We demonstrate that models that share a common origin, even with differences in scale or architecture, exhibit highly consistent patterns of neuronal activity within their functional networks across diverse input samples. In contrast, models trained independently on distinct data or with different objectives fail to preserve such activity alignment. Unlike conventional approaches, our method requires only a few samples for verification, preserves model utility, and remains robust to common model modifications (such as fine-tuning, pruning, and parameter permutation), as well as to comparisons across diverse architectures and dimensionalities. FNF thus provides model owners and third parties with a simple, non-invasive, and effective tool for protecting LLM intellectual property. The code is available at \url{https://github.com/WhatAboutMyStar/LLM_ACTIVATION}.
\end{abstract}


\section{Introduction}
Training large language models (LLMs) is extremely costly, requiring vast amounts of time and financial resources \cite{zhao2023survey,liu2023summary,WANG2024,LIU2025129190}. Although open-source models are typically released under specific licenses to protect intellectual property and prevent misuse or unauthorized commercial exploitation, some developers disregard these terms, publicly claiming to have trained “original” models from scratch, while in reality they have merely copied weights from existing models and applied superficial modifications such as minor fine-tuning or obfuscation. Therefore, extracting the unique fingerprint of a model is critical. Such fingerprints enable developers to reliably determine whether a suspect model is a derivative of a victim model or a genuinely original creation.

Traditional approaches typically rely on watermarking techniques \cite{peng2023you,xu2024instructional,xu2025copyright} that either invasively embed watermark signals directly into model parameters or train the model to exhibit anomalous prediction behavior in response to specific trigger phrases. However, such methods often degrade model performance and are inapplicable to models that have already been released \cite{russinovich2024hey}. Recently, fingerprinting methods based on intrinsic model properties have gained attention \cite{zhang2025reef}. Due to differences in training data, architecture, and optimization procedures, models from different families develop markedly distinct weight distributions and internal feature representations. These intrinsic differences enable non-invasive (effective) fingerprinting without requiring additional training or compromising model performance. 

However, existing methods focus primarily on comparing models with identical architectures and scales (or their derivatives) while overlooking a more subtle form of model reuse: structural expansion-based fine-tuning,  referred to as “weight repackaging.” For example, a developer can take an open-source LLM and add new layers, expand MLP dimensions, or make similar architectural enhancements, followed by additional training. Despite the altered scale and structure, the core weights remain largely inherited from the original model. Therefore, we need a fingerprinting method capable of comparing the similarity of LLMs with different hidden layer dimensions and architectures.

To address this challenge, we propose Functional Network Fingerprinting (FNF), a simple yet effective approach inspired by functional brain networks \cite{park2013structural,smith2009correspondence,shinn2023functional,qiang2024deep}. In neuroscience, the human brain exhibits stable patterns of functional connectivity, where groups of neurons consistently co-activate as integrated functional networks \cite{raichle2001default,smith2009correspondence,paus1997time,dong2020discovering}. LLM, as abstract computational analogues of biological neural systems, also have similar functional networks \cite{liu2025brain}: using methods analogous to those used in brain mapping, functional networks within LLMs that exhibit internally coherent activation patterns can be reliably identified. Leveraging this property, FNF can determine whether a suspect model is a derivative of a reference model or is independently trained by measuring the consistency of functional network activity in response to the same input stimuli.

This functional network fingerprint enables efficient identification of a model’s family affiliation. Our experiments reveal that models from the same family, such as Qwen-3B and Qwen-7B \cite{qwen2,yang2024qwen2}, exhibit strong cross-sample consistency in their functional network activity, whereas models from different families lack this property. Beyond enabling similarity analysis across models of different scales, our method exhibits strong robustness. Rather than relying on direct comparisons of parameters or static feature representations, FNF identifies consistency in functional network activity, a dynamic property defined by groups of co-activating neurons. Underlying co-activation patterns remain unchanged: rearranging weights does not alter which neurons jointly respond to a given input. Consequently, common obfuscation techniques such as parameter permutation, pruning, or other structural disguises have no effect on our ability to locate and compare functional networks. Even if the suspect model has repackaged the weights of the victim model and exhibits different hidden dimensions and architecture, it can still be identified by our method.

The proposed method FNF, a training-free method that leverages the intrinsic functional activity of LLMs to determine whether a suspect model is derived from a victim model. Unlike conventional approaches based on weight similarity or embedded watermarks, FNF preserves model performance and exhibits strong robustness against common obfuscation techniques. Moreover, it can effectively identify familial relationships across models of different scales and embedding dimensions within the same series. This provides an efficient and reliable means for model auditing and intellectual property protection.

\section{Related Works}
Existing model fingerprinting approaches fall into two main categories: invasive and non-invasive methods \cite{xu2025copyright}. Invasive methods fall into two types: one embeds watermarks or identifies information directly into model parameters, often using backdoor triggers \cite{peng2023you,adi2018turning,zhang2018protecting,li2019prove} or digital signatures \cite{guo2018watermarking}, and the other fine-tunes the model to produce specific behaviors in response to predefined trigger phrases \cite{li2023plmmark,xu2024instructional}. Both approaches have inherent limitations and inevitably degrade model performance.

In contrast, non-invasive methods extract identity-related information solely from the inherent properties of the model, without any modification to its structure or training process. Some methods perform post-hoc detection by analyzing model output \cite{mitchell2023detectgpt,sadasivan2023can,wu2023llmdet}, using semantic preferences or linguistic patterns to infer its origin \cite{iourovitski2024hide,pasquini2025llmmap,mcgovern2024your}. Others compare weight similarities \cite{chen2023deepjudge,chen2022copy,zeng2024huref}, but are easily defeated by simple parameter permutation. Both lack sufficient robustness. A third line of work, such as REEF \cite{zhang2025reef}, leverages LLM feature representations and offers improved robustness over traditional approaches. However, these methods ignore scenarios involving models with different depths, dimensions, or architectures, and thus fail to handle structural expansion-based fine-tuning or “weight repackaging,” significantly limiting their applicability.

Our FNF method captures consistent patterns in functional network activity (a dynamic property arising from co-activating groups of neurons) and remains robust against various model obfuscation techniques, effectively overcoming the limitations of traditional approaches.

\section{Preliminaries}
\textbf{Functional Brain Networks, FBNs:} Functional brain networks (FBNs), a concept from neuroscience \cite{bullmore2009complex,liu2023spatial,liu2024spatial,liu2024mapping,he2023multi,qiang2022learning,qiang2024deep}, refer to sets of brain regions that consistently co-activate. The human brain exhibits many such networks with stable activation patterns, which play key roles in cognitive functions, for example, the well-known default mode network (DMN) \cite{raichle2015brain,smith2009correspondence}. We bring the concept of functional brain networks to the LLM analysis. We treat neuron signals in LLMs as fMRI-like signals. We use methods like Independent Component Analysis (ICA), a standard tool in neuroscience to identify functional brain networks, to find functional networks in LLMs. Each LLM family has unique functional networks and activity patterns. By comparing the consistency of these functional activities, we can determine whether two models belong to the same family.

\textbf{Independent Component Analysis, ICA:} ICA is a common method for identifying functional networks. ICA assumes that the observed fMRI data is a linear mixture of statistically independent source signals. Its goal is to "unmix" the data and recover these underlying independent components, each corresponding to a spatial pattern that represents a functional network. These spatial patterns may be associated with specific brain functions. The success of ICA in fMRI analysis offers a promising framework to identify functional networks in LLMs. Just as fMRI signals capture the collective response of biological neurons to stimuli, neuron signals in LLMs reflect changes in internal state during input processing. By treating LLM signals as analogous to fMRI signals, we can directly apply ICA to extract independent functional components from the model. In this work, the neuron signals from the output of each Transformer block serve as the input to ICA. We use a spatial ICA variant, called canonical ICA (CanICA) \cite{varoquaux2010group,varoquaux2010ica}, to identify functional networks. It first reduces the dimensionality of the data using PCA and then applies FastICA \cite{hyvarinen2000independent} to the low-dimensional representation to recover independent spatial components. 

\section{Method}
\subsection{Functional Activation Consistency}
When different human individuals receive the same task stimulus, their fMRI signals show similar activation patterns in certain regions of the brain \cite{liu2024spatial,CALHOUN20011080,calhoun2008modulation,liu2024mapping,liu2023spatial,he2023multi}. This inter-individual consistency arises because humans share a common neuroanatomical blueprint shaped by evolution. However, for different series of LLMs should not be viewed as analogous to individual humans within a single species. Instead, they are better conceptualized as distinct “neural species”, each shaped by its own evolutionary trajectory defined by architecture choices, tokenizer design, training data and optimization protocols. For instance, humans and non-human primates (e.g., macaques) show strikingly similar engagement of homologous cortical regions when performing cognitive tasks \cite{lu2012rat}, reflecting shared neuroanatomical and functional architectures shaped by common ancestry. In contrast, more distantly related species exhibit markedly different activation pattern for analogous behaviors.

Thus, analogous to the organization of biological brains, LLMs derived from the same training lineage exhibit consistent activation patterns across the corresponding functional networks in response to identical inputs. In contrast, models from different lineages lack this alignment, showing divergent functional activations. We formalize this phenomenon as Functional Activation Consistency (FAC) and use it to construct a model lineage fingerprint.

\subsection{Extract Functional Networks in LLMs}
In the human brain, functional networks are typically groups of neurons that co-activate during specific cognitive tasks \cite{smith2009correspondence,paus1997time,raichle2001default,dong2020discovering}. Similarly, LLMs produce sets of co-activated neurons in response to the same input \cite{liu2025brain}. These functional networks can be decomposed using ICA; in this work, we apply CanICA to neuron signals from the output of each Transformer block in the LLM. 

Consider $N$ input sequences processed by an LLM. For each sequence $n = 1, 2, \dots, N$, we extract the neuron signals from the output of a specific Transformer block, yielding a matrix: $\mathbf{X}^{(n)} \in \mathbb{R}^{T_n \times D}$, where $T_n$ is the sequence length (number of tokens) for the $n$-th input, $D$ is the dimensionality of the Transformer block's output (i.e., number of neurons or hidden units).

CanICA is a group-wise analysis method for identifying functional networks. The concatenated multi-sample signal matrix is: 

\begin{equation}
\mathbf{X} = 
\begin{bmatrix}
\mathbf{X}^{(1)}
\dots
\mathbf{X}^{(N)}
\end{bmatrix}
\in \mathbb{R}^{T \times D}, T = \sum_{n=1}^N T_n.
\end{equation}

To reduce dimensionality and remove second-order correlations, CanICA apply PCA whitening: 
\begin{equation}
    \mathbf{Z} = \mathbf{X} \mathbf{W} \in \mathbb{R}^{T \times K}
\end{equation}

Where $K \ll D$ is the number of target components (latent functional networks), $\mathbf{W} \in \mathbb{R}^{D \times K}$ contains the top $K$ right singular vectors scaled by inverse singular values (whitening transform), $\mathbf{Z}$ is the whitened data. Each row $\mathbf{z}_t \in \mathbb{R}^K$ represents the $t$-th token’s neural state in a low-dimensional, decorrelated, and variance-normalized latent space. The whitening ensures: $\frac{1}{T} \mathbf{Z}^\top \mathbf{Z} = \mathbf{I}_K$, so that all dimensions of $\mathbf{Z}$ are uncorrelated and have unit variance.

CanICA model the whitened data as a linear mixture of spatially independent components: $\mathbf{Z} = \mathbf{S} \mathbf{A}^\top$, where $\mathbf{S} \in \mathbb{R}^{T \times K}$ refer to time (token) courses, dynamic activation patterns across tokens; $\mathbf{A} \in \mathbb{R}^{D \times K}$ refer to spatial maps, each column $\mathbf{a}_k \in \mathbb{R}^D$ represents a functional network. The final result we need in CanICA is the functional network in $\mathbf{A}$.

\begin{figure*}
    \vskip 0.2in
    \centering
    \includegraphics[width=\linewidth]{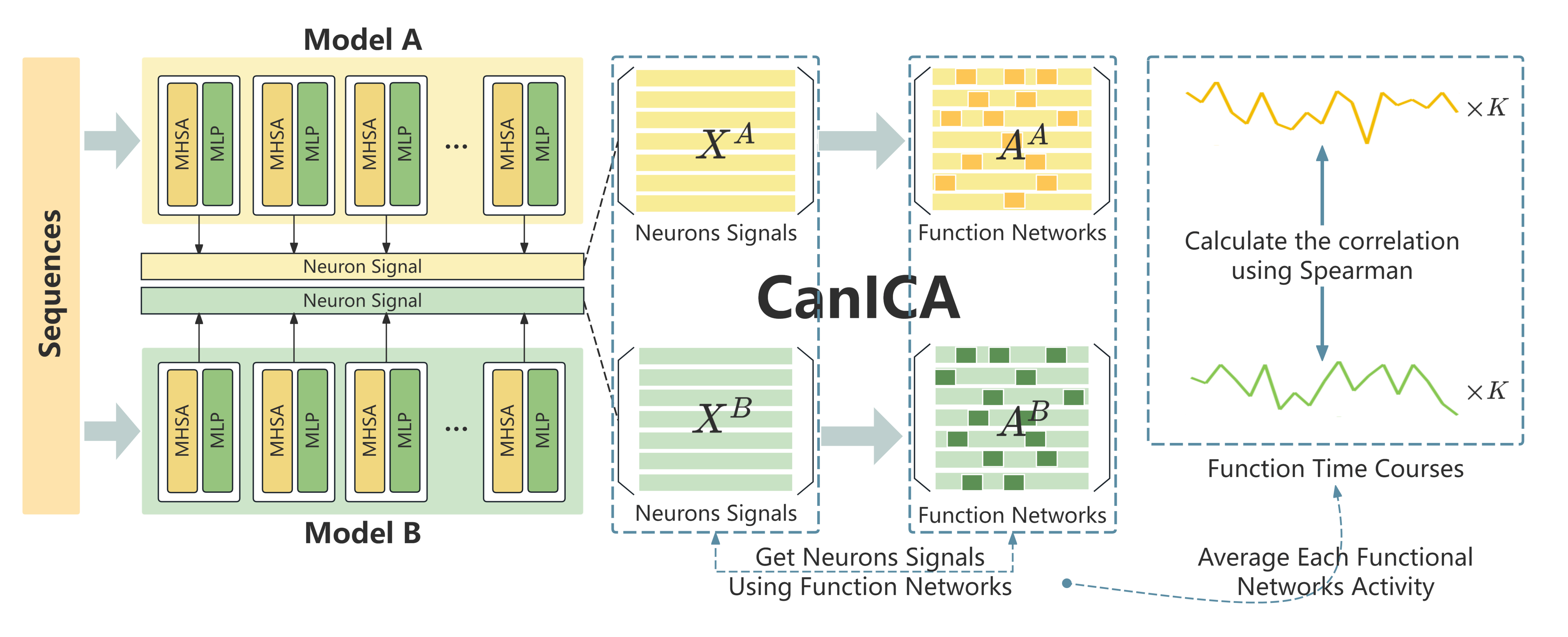}
    \caption{The total framework of Functional Network Fingerprint.}
    \label{fig:framework}
\end{figure*}

\subsection{Functional Network Fingerprint}
The pipline of the method is shown in Figure~\ref{fig:framework}. Select two LLMs for comparison: Model A and Model B. Process the same set of $N$ input sequences through the 2 models to obtain: $\mathbf{X}^{\text{A},(n)} \in \mathbb{R}^{T_n \times D}$ and $\mathbf{X}^{\text{B},(n)} \in \mathbb{R}^{T_n \times D}, n = 1,\dots,N$
where $T_n$ is the sequence length (number of tokens) of the $n$-th input, and $D$ is the hidden dimension (number of neurons). 

We apply CanICA independently to each model’s concatenated data to extract $K$ functional networks. For Model A: obtain spatial maps $\mathbf{A}^\text{A} = [\mathbf{a}_1^\text{A}, \dots, \mathbf{a}_K^\text{A}] \in \mathbb{R}^{D \times K}$. For Model B: obtain spatial maps $\mathbf{A}^\text{B} = [\mathbf{a}_1^\text{B}, \dots, \mathbf{a}_K^\text{B}] \in \mathbb{R}^{D \times K}$. Where each column $\mathbf{a}_k$ represents a functional network. Following standard practice in neuroimaging, we threshold each spatial map using a conventional fixed cutoff to identify the most salient neurons, yielding binary masks that select the core features of each network.

For each sample $n$, component $i$ (Model A), and component $j$ (Model B), we compute the mean neuron activities over the masked neurons to obtain 1D time courses: $\mathbf{s}_k^{\text{A},(n)} \in \mathbb{R}^{T_n}$ and $\mathbf{s}_k^{\text{B},(n)} \in \mathbb{R}^{T_n}$. This yields $K$ functional activity sequences per model per sample. 

We then compute the Spearman rank correlation between all pairs of functional time courses:
\begin{equation}
    \rho_{ij}^{(n)} = \operatorname{Spearman}\left( \mathbf{s}_i^{\text{A},(n)},\ \mathbf{s}_j^{\text{B},(n)} \right), \quad \forall\, i,j \in \{1,\dots,K\}.
\end{equation}

This yields $N$ correlation values for the network pair $(i,j)$, each input sample having one. We then compute the average consistency across all samples:
\begin{equation}
    \bar{\rho}_{ij} = \frac{1}{N} \sum_{n=1}^{N} \rho_{ij}^{(n)}
\end{equation}

The resulting $K \times K$ matrix $\bar{\mathbf{R}} = [\bar{\rho}_{ij}]$ quantifies the average dynamic similarity between all pairs of functional networks between the two models. The overall pipeline is summarized in Algorithm~\ref{alg:fnf}.
\begin{algorithm}[t]
\caption{Functional Network Fingerprint (FNF)}
\label{alg:fnf}
\begin{algorithmic}[1]
\REQUIRE Activations from two models: $\mathbf{X}^\text{A}, \mathbf{X}^\text{B} \in \mathbb{R}^{T \times D}$; \\
         Number of functional networks $K$.
\ENSURE Cross-model similarity matrix $\bar{\mathbf{R}} \in \mathbb{R}^{K \times K}$.

\STATE // Extract spatial functional networks via CanICA
\STATE $\mathbf{A}^\text{A} \gets \textsc{CanICA}(\mathbf{X}^\text{A}, K)$ \COMMENT{$\mathbf{A}^\text{A} \in \mathbb{R}^{D \times K}$}
\STATE $\mathbf{A}^\text{B} \gets \textsc{CanICA}(\mathbf{X}^\text{B}, K)$

\STATE // Compute functional time courses for all $N$ samples
\FOR{$n = 1$, \dots, $N$}
    \FOR{$i = 1$, \dots, $K$}
        \STATE $\mathbf{s}_i^{\text{A},(n)} \gets \text{mean of } \mathbf{X}^{\text{A},(n)} \text{ over neurons in } \mathbf{a}_i^\text{A}$
        \STATE $\mathbf{s}_i^{\text{B},(n)} \gets \text{mean of } \mathbf{X}^{\text{B},(n)} \text{ over neurons in } \mathbf{a}_i^\text{B}$
    \ENDFOR
\ENDFOR

\STATE // Compute average dynamic similarity
\FOR{$i = 1, \dots, K$}
    \FOR{$j = 1, \dots, K$}
        \STATE $\rho_{ij}^{(n)} \gets \operatorname{Spearman}\big( \mathbf{s}_i^{\text{A},(n)},\, \mathbf{s}_j^{\text{B},(n)} \big), \quad \forall n$
        \STATE $\bar{\rho}_{ij} \gets \frac{1}{N} \sum_{n=1}^N \rho_{ij}^{(n)}$
    \ENDFOR
\ENDFOR

\STATE $\bar{\mathbf{R}} = [\bar{\rho}_{ij}]$
\end{algorithmic}
\end{algorithm}

Importantly, we used the Spearman rank correlation rather than the Pearson correlation because our primary interest lies in the consistency of functional activity trends. Pearson correlation, being sensitive to linear relationships and absolute values, can yield spuriously high scores for time courses that appear numerically similar but follow different dynamic trajectories (e.g., one monotonically rising while the other peaks early). In contrast, the Spearman correlation operates on rank orders, making it robust to monotonic transformations and better aligned with our goal of assessing qualitative trend consistency. Although no scalar metric fully captures complex temporal alignment, Spearman correlation currently offers the best trade-off between interpretability, robustness, and sensitivity to meaningful functional similarity in this context.

Notably, the Spearman correlation coefficient serves as an indicator of functional alignment rather than a definitive binary criterion. We consider two models to share a common origin only when their corresponding functional networks show moderate-to-strong rank-order consistency across the majority of input samples. This cross-sample stability ensures that the similarity reflects a systematic functional correspondence, not a spurious or instance-specific match.

\section{Results}
\subsection{Experiments Setting}
We use WikiText-2 \cite{merity2016pointer} to evaluate model similarity. Its samples typically span hundreds of tokens, providing sufficient context for reliable fingerprinting. We analyze functional similarity across a broad range of models to capture the effects of architecture, scale, training stage, data, and generational updates. 

In our experiments, we used 10 samples from Wikitext-2 as input to CanICA for group-level analysis, decomposing the data into 64 functional networks (i.e., $K = 64$ ).

This setup enables a systematic evaluation of how model similarity changes under various design and training factors.

\subsection{Evaluation}

\begin{table*}[t]
    \caption{The FNF between pre-trained LLaMA2-7B and Qwen2.5-7B models with their fine-tuned model. The FNF shows the largest value in $K \times K$ matrix $\bar{\mathbf{R}}$. }
    \label{tab::pretrain_finetune}
    \begin{center}
        \begin{tabular}{ccc}
            \toprule
             & LLaMA2-7B-hf and LLaMA2-7B-chat-hf & Qwen2.5-7B and Qwen2.5-7B-Instruct\\
            \midrule
            CKA & 0.9877 & 0.9225 \\
            FNF &  0.9493   &   0.9582 \\
            \bottomrule
        \end{tabular}
    \end{center}
    \vskip -0.1in
\end{table*}

\begin{table*}
    \caption{The FNF and CKA between LLaMA2-7B-chat and LLaMA2 fine-tuned variants trained on different amounts of data. The FNF shows the largest value in $K \times K$ matrix $\bar{\mathbf{R}}$. }
    \label{tab::llama_finetune}
    \begin{center}
        \begin{tabular}{ccccc}
            \toprule
            LLaMA2-7B-chat & Vicuna-v1.5 & Chinesellama-2-7B & Codellama-7B & Llemma-7B\\
            Tokens &   370M   & 13B  &  500B & 700B \\
            \midrule
             CKA & 0.7396 & 0.9983  & 0.8833   & 0.9701\\
             FNF & 0.5963 & 0.9642  &  0.9184  &  0.8326\\
            \bottomrule
        \end{tabular}
    \end{center}
    \vskip -0.1in
\end{table*}
\begin{table}
    \caption{The FNF and CKA between LLaMA2-7B-chat and the models share an identical architecture but are trained on different datasets from scratch. The FNF shows the largest value in $K \times K$ matrix $\bar{\mathbf{R}}$. }
    \label{tab::train-data-variants}
    \begin{center}
        \begin{tabular}{ccc}
            \toprule
            LLaMA2-7B-chat & Amber & Openllama-7B\\
            \midrule
             CKA & 0.3673 & 0.4159  \\
             FNF & 0.3015 & 0.2061 \\
            \bottomrule
        \end{tabular}
    \end{center}
    \vskip -0.1in
\end{table}

\begin{table*}
    \caption{The FNF and CKA between LLaMA2-7B-Chat and Vicuna-v1.5-7B and their pruned models, covering both structured pruning (width and depth pruning) and unstructured pruning. The FNF shows the largest value in $K \times K$ matrix $\bar{\mathbf{R}}$. }
    \label{tab::pruning}
    \begin{center}
        \begin{tabular}{ccccc}
            \toprule
            Model & FLAP & Wanda & Shortened LLaMA & Wanda-sp\\
            & (Width Pruning) & (Unstructured) & (Depth Pruning) & (Width Pruning)\\
            \midrule
            LLaMA2 (CKA) & 0.8236  & 0.9991 & 0.9996 &  \textbf{0.2767}\\
            Vicuna (CKA) & 0.9863  & 0.9107 & 0.9100 &  \textbf{0.0012} \\
            LLaMA2 (FNF) & 0.6172  & 0.9451 & 0.5965 & 0.5007 \\
            Vicuna (FNF) & 0.9097  & 0.9249 & 0.9692 & 0.6751  \\
            \bottomrule
        \end{tabular}
    \end{center}
    \vskip -0.1in
\end{table*}

\begin{table}
    \caption{The FNF and CKA between Evollm-jp-7b and its merged models (Shisa-gamma-7b-v1, WizardMath-7b-1.1, and Abel-7b-002). The FNF shows the largest value in $K \times K$ matrix $\bar{\mathbf{R}}$. }
    \label{tab::merging}
    \begin{center}
    \resizebox{\columnwidth}{!}{
        \begin{tabular}{ccccc}
            \toprule
            &  Shisa-gamma-7b-v1 &  WizardMath-7b-1.1 & Abel-7b-002 \\ 
            \midrule
             CKA & 0.9635  & 0.9526  &  0.9374 \\
             FNF & 0.9595 &  0.8442  &  0.9589 \\
            \bottomrule
        \end{tabular}
        }
    \end{center}
    \vskip -0.1in
\end{table}

\begin{table}
    \caption{The Permutation experiment results (FNF and CKA) between LLaMA2-7B, Vicuna-v1.5-7B, and ChatGLM3-6B. The FNF shows the largest value in $K \times K$ matrix $\bar{\mathbf{R}}$. }
    \label{tab::permutation}
    \begin{center}
    \resizebox{\columnwidth}{!}{
        \begin{tabular}{ccccc}
            \toprule
            & LLaMA2-7B-chat & Vicuna-v1.5-7B & ChatGLM3-6B \\ 
            \midrule
             CKA & 1.0000  & 1.0000  &  1.0000 \\
             FNF & 0.9986  & 0.9972  &  0.9971  \\
            \bottomrule
        \end{tabular}
        }
    \end{center}
    \vskip -0.1in
\end{table}

\subsubsection{Baseline Method}
The baseline method compared in this paper is REEF \cite{zhang2025reef}, which computes the Centered Kernel Alignment (CKA) between the internal representations of a victim model and a suspect model to determine whether they share a common origin. REEF is currently the only publicly available method capable of measuring similarity between models with mismatched input feature dimensions, enabling cross-architecture comparisons. This capability makes REEF particularly promising for detecting model similarity even when models have undergone weight repackaging (such as through dimension expansion or adding new layers) across different architectures. Moreover, REEF is currently the best-performing method among internal representation–based fingerprinting approaches and, compared to other baselines, can effectively overcome common model obfuscation techniques.

\subsubsection{Fine-tuning}
Extensive fine-tuning often leads to significant changes in model parameters, which can undermine the reliability of weight-based fingerprinting methods. We first used FNF to compare the pre-trained LLaMA2-7B \cite{touvron2023llama} and Qwen2.5-7B models with their fine-tuned model. The results are shown in Table~\ref{tab::pretrain_finetune}. The table shows that some functional networks exhibit consistent activity between the pre-trained and fine-tuned versions of LLaMA-2 and Qwen2.5. To better illustrate the consistency of functional network activity, Figure~\ref{fig:fna} shows the most similar functional network activities between LLaMA2-7B and its fine-tuned model, LLaMA2-7B-chat.
\begin{figure}[h]
    \vskip 0.2in
    \centering
    \includegraphics[width=\linewidth]{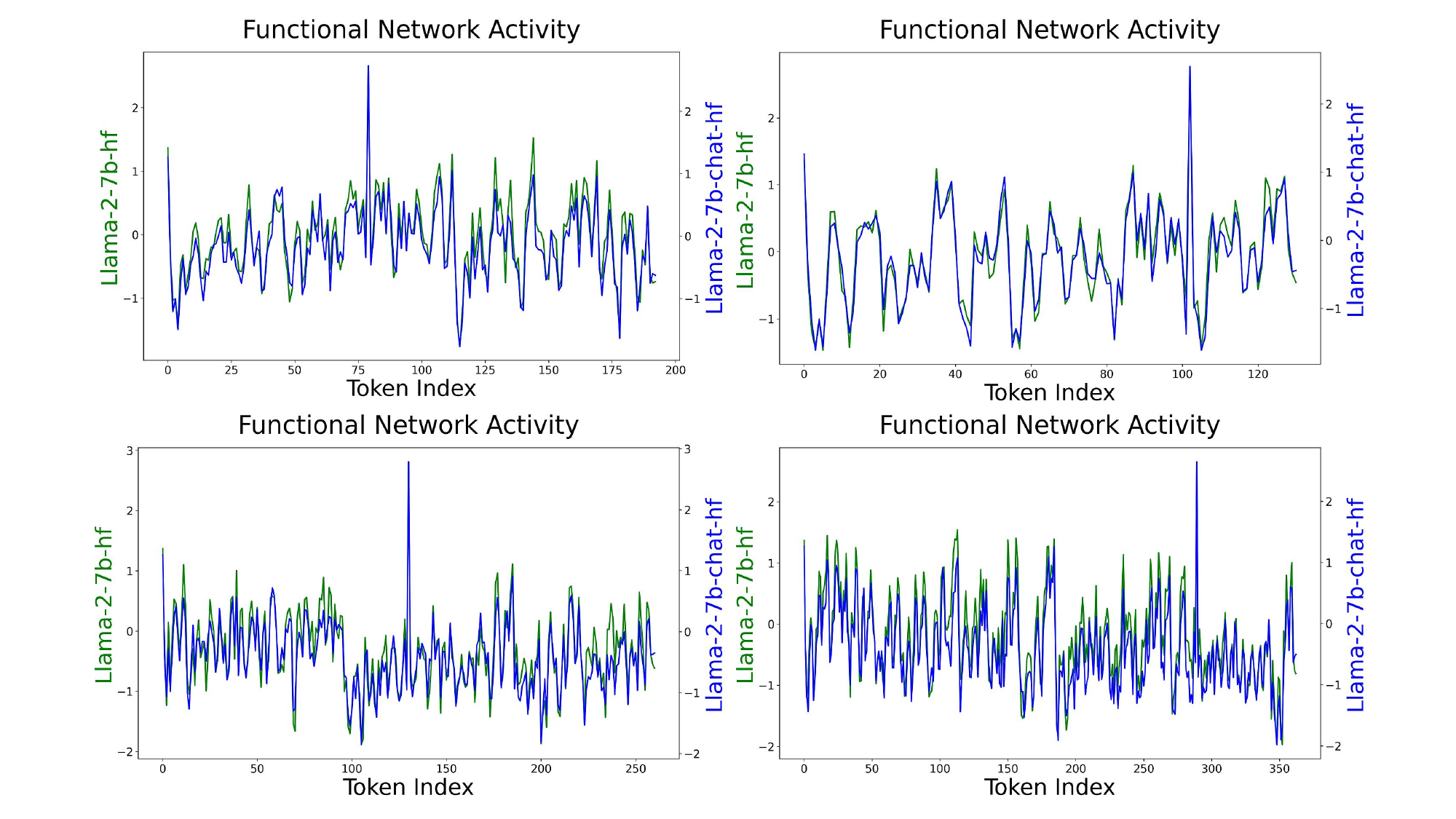}
    \caption{Functional network activity of LLaMA2-7B-hf and LLaMA2-7B-chat-hf across different samples.}
    \label{fig:fna}
\end{figure}

\begin{figure}[h]
    \vskip 0.2in
    \centering
    \includegraphics[width=\linewidth]{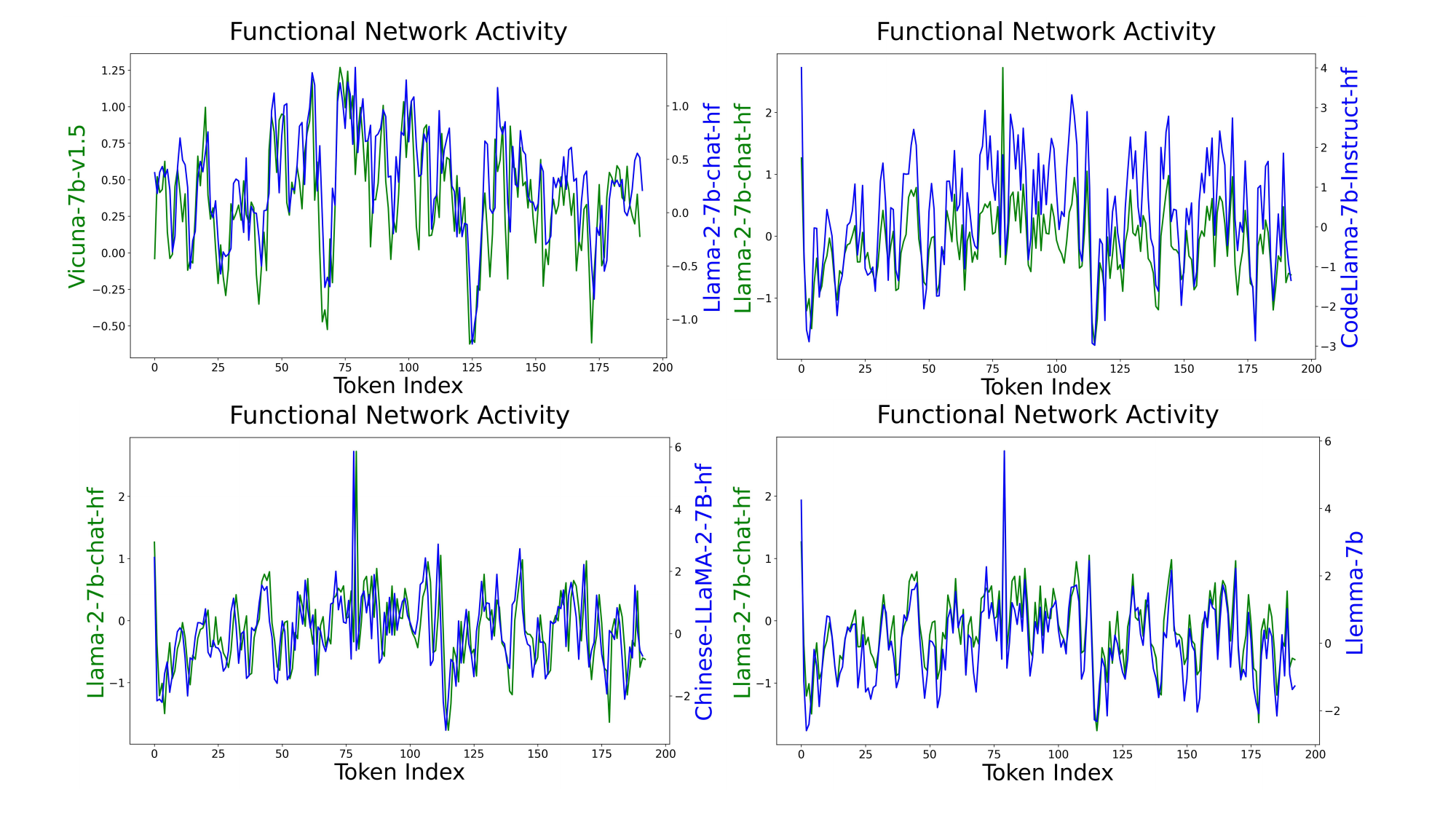}
    \caption{Functional network activity of LLaMA2-7B-chat-hf and fine-tuned variants trained on different amounts of data. }
    \label{fig:fna}
\end{figure}

\begin{figure}[h]
    \vskip 0.2in
    \centering
    \includegraphics[width=\linewidth]{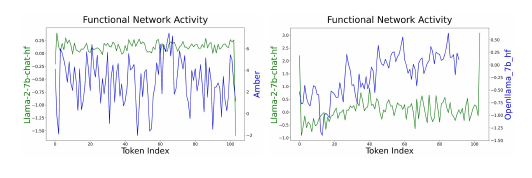}
    \caption{Examples of functional network activity from Amber and OpenLLaMA-7B.}
    \label{fig:amber-openllama}
\end{figure}

We then further examined whether the FNFs remain consistent between LLaMA2-7B-chat and models fine-tuned from LLaMA2 using varying amounts of training data \cite{chiang2023vicuna,azerbayev2023llemma}. The results are shown in Table~\ref{tab::llama_finetune}. We observe a reduction in the FNF score, but this does not undermine its validity as a fingerprint: model relatedness is ultimately determined by the consistency of functional network activity, measured via Spearman correlation. Correlations above 0.5 are considered strong and, as shown in Figure~\ref{tab::llama_finetune}, the consistency of functional network activity is clearly evident. Even after fine-tuning on 700B tokens (a process that incurs substantial computational cost) the fingerprint of the original model remains clearly detectable, demonstrating the robustness of our method.

\subsubsection{Training-data Variants}
We consider a set of models that share an identical architecture but are trained on different datasets from scratch, referred to as training-data variants. Taking LLaMA2 as an example, OpenLLaMA-7B \cite{openlm2023openllama} and Amber \cite{liu2023llm360} are models trained on different datasets using the LLaMA2 architecture and are thus theoretically unrelated to LLaMA2. A model fingerprinting method should be able to distinguish between them. 

The results are shown in Table~\ref{tab::train-data-variants}. We observe that LLaMA2-7B shows weak Spearman correlations with both Amber and OpenLLaMA. Athough Amber’s correlation is relatively higher, reaching the threshold for weak correlation, some individual samples do not exhibit correspondence. In Figure~\ref{fig:amber-openllama}, show the functional network activities of Amber and OpenLLaMA. Their activity patterns show little alignment with LLaMA2, reflecting low actual correlation, consistent with the fact that both Amber and OpenLLaMA are LLMs trained from scratch on different data and are not derived from LLaMA2.

\subsubsection{Model Pruning}
Model pruning is a common model compression technique that removes a portion of a model’s neurons, which can significantly alter the integrity of its functional structure. In our experiments, we treat the LLaMA2-7B model as the victim and consider its pruned variants (produced by various pruning methods) as suspect models. We evaluate multiple pruning strategies, including structured and unstructured pruning \cite{sun2023simple}, as well as depth pruning(layer removal \cite{kim2024shortened}) and width pruning (neuron/channel removal \cite{an2024fluctuation}) approaches.

We conducted pruning experiments in LLaMA2-7B-chat and Vicuna-v1.5-7B to assess the impact of pruning on our method. In our experiments, we pruned the models using FLAP \cite{an2024fluctuation}(width pruning), Shortened LLM (depth pruning) \cite{kim2024shortened}, Wanda (unstructured pruning) \cite{sun2023simple}, and its structured variant Wanda-sp \cite{an2024fluctuation}, then compared the similarity between each pruned model and its original model. We pruned the models, removing 50\% of their parameters. The results are shown in Table~\ref{tab::pruning}. The pruned models retain highly consistent functional network activity with their original models, exhibiting strong correlation, demonstrating that our method can detect pruning-based model obfuscation. We can see that even after 50\% pruning, our method still correctly identifies the models as sharing a common origin. In contrast, REEF fails to distinguish them as related under heavy pruning, specifically, it cannot recognize their common origin at the 50\% pruning rate with Wanda-sp. Excessive pruning combined with suboptimal pruning strategies can impair model functionality. After pruning with Wanda-sp, the consistency of functional network activity measured by our FNF method also decreases, though it remains in the strong correlation range.

\begin{table}
    \caption{The Scaling Transformation experiment results (FNF and CKA) between LLaMA2-7B, Qwen2.5-7B-Instruct, and ChatGLM3-6B. The FNF shows the largest value in $K \times K$ matrix $\bar{\mathbf{R}}$. }
    \label{tab::scaling}
    \begin{center}
    \resizebox{\columnwidth}{!}{
        \begin{tabular}{ccccc}
            \toprule
            & LLaMA2-7B-chat & Qwen2.5-7B-Instruct & ChatGLM3-6B \\ 
            \midrule
             CKA & 1.0000  & 1.0000  &  1.0000 \\
             FNF & 0.9984  & 0.9832  &  0.9999  \\
            \bottomrule
        \end{tabular}
        }
    \end{center}
    \vskip -0.1in
\end{table}

\begin{table*}[t]
    \caption{The FNF and CKA between different models. The FNF shows the largest value in $K \times K$ matrix $\bar{\mathbf{R}}$}.
    \label{tab::cross}
    \begin{center}
        \begin{tabular}{ccccccc}
            \toprule
            Qwen2.5-7B-Instruct& Qwen1.5-7B-Instruct & Qwen2.5-3B-Instruct & Qwen2.5-1.5B-Instruct \\ 
            \midrule
             CKA & 0.7958  & 0.7988  &  0.6832 \\
             FNF & 0.5752  & 0.5369  &  0.4648  \\
            \midrule
            Qwen2.5-7B-Instruct & LLaMA3-8B & ChatGLM3-6B & GLM4-9B \\
            \midrule
            CKA & \textbf{0.7944}  & 0.6842  &  \textbf{0.8567} \\
            FNF & 0.3765  & 0.1732  &  0.2560  \\
            \midrule
            ChatGLM2-6B & LLaMA3-8B & ChatGLM3-6B & GLM4-9B \\
            \midrule
            CKA & \textbf{0.9071}  & 0.7475  &  \textbf{0.8845} \\
            FNF & 0.3221  & 0.5799  &  0.2897  \\
            \bottomrule
        \end{tabular}
    \end{center}
    \vskip -0.1in
\end{table*}
\subsubsection{Model Merging}
Model merging technology for LLMs has emerged as a highly active research area in recent years \cite{yang2024model}. Its core idea is to combine multiple pre-trained or fine-tuned LLMs into a single model (without additional training) so that the resulting model inherits complementary strengths from its constituents, such as enhanced multitask performance, domain adaptability, or robustness. Because model merging integrates parameters from multiple distinct LLMs, an effective model fingerprinting method must be able to identify all potential suspect models as well as their corresponding victim models.

We selected Evollm-jp-7b \cite{akiba2025evolutionary} as our experimental model, which fuses three models of identical architecture (Shisa-gamma-7b-v1, WizardMath-7b-1.1, and Abel-7b-002) using weighted parameters. The results are shown in Table~\ref{tab::merging}. Our method effectively captures the consistency in functional network activity among the merged models, successfully identifying the model's origin. The fusion of model parameters inherently aligns with the ICA core assumption that observed signals are linear mixtures of independent source components. Because different models’ functional signals are combined in a linear manner, ICA can effectively disentangle them into distinct functional networks. This principled separation explains why the recovered functional time courses exhibit strong cross-model correlations.

\begin{table}
    \caption{The weight repackaging experiment results (FNF and CKA) between solar-10.7B, Mistral-7B and Mistral-11B. The FNF shows the largest value in $K \times K$ matrix $\bar{\mathbf{R}}$. }
    \label{tab::repackaging}
    \begin{center}
    \resizebox{\columnwidth}{!}{
        \begin{tabular}{lcc}
            \toprule
            & CKA & FNF  \\ 
            \midrule
             Mistral-7B \& Solar-10.7B &  0.3226 & 0.5743 \\
             Mistral-7B \& Mistral-11B &  0.9923 & 0.5820 \\
             Solar-10.7B \& Mistral-11B & 0.3320& 0.7958 \\
            \bottomrule
        \end{tabular}
        }
    \end{center}
    \vskip -0.1in
\end{table}

\subsubsection{Permutation and Scaling Transformation}
Permuting the parameters and Scaling transformation of an LLM can effectively disguise the model without altering its architecture or affecting its outputs, allowing it to evade simple fingerprinting techniques such as those based on direct weight comparison \cite{fernandez2024functional}. 

We conducted experiments on LLaMA2-7B-chat, Vicuna-v1.5-7B, Qwen2.5-7B, and ChatGLM3-6B-base \cite{glm2024chatglm}. The scaling transformation experiments were conducted after permutation. The results are shown in Table~\ref{tab::permutation} and Table~\ref{tab::scaling}. Our method successfully captures their shared origin.

\subsubsection{Cross-model and Weight Repackaging}
We evaluate the robustness of our fingerprint under cross-model comparison. A practical form of model plagiarism involves copying weights from a victim model and then altering its architecture—for example, by expanding embedding dimensions, adding or removing layers, or applying other structural modifications before further training. Crucially, such attacks deliberately change the model’s architecture and parameter dimensions, rendering most existing fingerprinting methods inapplicable, as they assume fixed or identical model structures. This limitation highlights the urgent need to first investigate whether a fingerprinting approach can reliably compare models with heterogeneous architectures and dimensionalities. To address this gap, we conduct comprehensive evaluations across distinct model families as well as scaled variants within the same family (e.g., models with different parameter counts), demonstrating that our method remains effective despite significant architectural and dimensional differences.

We compare our proposed FNF method with CKA. As shown in Table~\ref{tab::cross}, CKA consistently reports high similarity scores even between models that are clearly unrelated (e.g., Qwen2.5-7B-Instruct vs. GLM4-9B: CKA = 0.8567 and ChatGLM2-6B vs. LLaMA3-8B: CKA=0.9071). In contrast, FNF yields substantially lower scores for such cross-family comparisons. Notably, when comparing models within the same family (e.g., Qwen2.5-7B-Instruct vs. Qwen1.5-7B-Instruct), FNF still reports moderate-to-strong consistency (0.5752), aligning with their shared lineage, while maintaining clear separation from out-of-family models. ChatGLM2 and ChatGLM3 are more similar to each other, while both are significantly different from GLM4. This is consistent with their actual development history. GLM4 is a completely new version in the GLM series, with a different architecture and different training methods. In contrast, ChatGLM2 and ChatGLM3 share identical architectures, differing only in parts of their training data.

The model solar-10.7b \cite{kim2024solar} is derived from the weights of Mistral-7B \cite{jiang2023mistral7b} by expanding its depth from 32 to 48 layers. Similarly, a community-released model, Mistral-11B (available at \url{https://huggingface.co/Undi95/Mistral-11B-v0.1}), applies the same type of architectural expansion. Both models align with our definition of weight repackaging (modifying the model structure while reusing the original weights). We evaluate these models using our method and CKA; as shown in Table~\ref{tab::repackaging}, our approach successfully identifies their common origin, whereas CKA fails to do so.


\section{Disussion and Conclusion}
Inspired by neuroscience research on functional brain networks, this paper proposes a model fingerprinting method for LLMs based on their functional networks. Our approach is data-driven and robust against various model obfuscation techniques, such as weight permutation, scaling transformation, model pruning, or architectural modifications while enabling comparisons across different models, architectures, and model scales. This paper is the first to propose the weight repackaging attack and validates the effectiveness of our method against this type of attack. Experiments demonstrate that LLMs exhibit distinct patterns of functional activity: the activation profiles of their functional networks serve as highly discriminative and reliable fingerprints. Our method is computationally efficient, easy to use, and highly practical, making it well-suited for protecting model intellectual property. Moreover, our method is highly interpretable: ICA decomposes the activations into multiple functional networks, and similar models exhibit alignment at the functional network level. Our approach holds significant potential to advance model interpretability more broadly.








\section*{Impact Statement}
This paper presents work whose goal is to advance the field of Machine Learning. There are many potential societal consequences of our work, none of which we feel must be specifically highlighted here.

\nocite{langley00}

\bibliography{example_paper}
\bibliographystyle{icml2026}

\newpage
\appendix
\onecolumn
\section{Appendix}

\subsection{Ablation Study}
The ablation study aims to evaluate the robustness of our method under varying numbers of input samples, different choices of \( K \) (the number of functional networks), and different input datasets.

We conducted experiments using ChatGLM2-6B and ChatGLM3-6B, first increasing the number of input samples from 10 to 100 while keeping \( K = 64 \). Among the 64 functional networks, the highest FNF value is 0.5778, which shows no significant difference from the result obtained with 10 samples (0.5799). This shows that the FNF method can reliably determine model similarity with only a small number of samples, resulting in a very low computational overhead.

In neuroscience, there is no universally correct value for \( K \); it depends on various factors and is typically chosen based on empirical experience. For our purpose, we only need to extract enough functional networks to reliably determine whether two models share a common origin. In this work, we set \( K = 64 \), but other values yield consistent conclusions. We present the FNF scores between ChatGLM2-6B and ChatGLM3-6B under different choices of \( K \) shown in Table~\ref{tab::n_componenst} to demonstrate this robustness. We observe that \( K \) should not be set too small or unnecessarily large, values around 64 are sufficient to reliably identify meaningful functional networks.

\begin{table}[h]
    \caption{The FNF score under different choices of \( K \).}
    \label{tab::n_componenst}
    \begin{center}
        \begin{tabular}{cccccc}
            \toprule
            K (Number of components)& 10 & 20 & 40 & 64 & 128\\ 
            \midrule
            ChatGLM2 vs ChatGLM3 & 0.3201&  0.3844 & 0.6202 &  0.5799 & 0.5101\\ 
            Qwen2.5-7B vs LLaMA3-8B&  0.1970 &  0.3069 & 0.3042 & 0.3765  & 0.3442\\ 
            \bottomrule
        \end{tabular}
    \end{center}
    \vskip -0.1in
\end{table}

We also conducted experiments on Wikitext-2, SST-2, SQuAD and a generated random token datasets using LLaMA2-7B-chat and Llemma-7B to assess the impact of different input datasets. As shown in Table~\ref{tab::dataset}, the performance in SST-2 is weaker, although it still shows a weak correlation, while the results in Wikitext-2, SQuAD and the generated random token dataset are consistently strong. Upon inspection, we found that SST-2 samples contain very few tokens (around 10 per sample), which limits the reliability of correlation estimation. In contrast, both SQuAD and Wikitext-2 provide much longer sequences (typically more than 100 tokens per sample). Each sample in the randomly generated token dataset contains more than 100 tokens, though fewer than those in SQuAD and Wikitext-2. Notably, even when the input consists of random, meaningless tokens, our FNF method can still identify functionally consistent networks, as long as each sample provides a sufficient number of tokens. In general, the key requirement is that input samples contain sufficient tokens to ensure statistically meaningful correlation estimates.

\begin{table}[h]
    \caption{The FNF score in different datasets. Each sample in the generated random token dataset contains more than 100 tokens, but less than SQuAD and Wikitext-2.}
    \label{tab::dataset}
    \begin{center}
        \begin{tabular}{cccccc}
            \toprule
            Dataset & Wikitext-2 & SQuAD & SST-2 & Random data\\ 
            \midrule
            LLaMA2-7B-chat vs Llemma-7B & 0.8718 & 0.8514 & 0.4353 & 0.7426\\ 
            \bottomrule
        \end{tabular}
    \end{center}
    \vskip -0.1in
\end{table}

\subsection{Overlap of Functional Networks in LLMs}

Models that share a common origin exhibit highly consistent functional network activity. Among models with the same architecture and scale, the spatial locations and overlap of functional networks are directly comparable. Does temporal consistency in functional activity also imply strong spatial alignment of these networks? To investigate whether functional networks are also aligned in spatial organization, we compute the spatial similarity between the corresponding functional networks of two models using the Intersection over Union (IoU). Specifically, for each functional network derived from Model A and its counterpart from Model B (both represented as binary masks over the \(D\)-dimensional hidden space) we calculate IoU as:

\begin{equation}
    \text{IoU} = \frac{|\mathcal{M}^{\text{A}} \cap \mathcal{M}^{\text{B}}|}{|\mathcal{M}^{\text{A}} \cup \mathcal{M}^{\text{B}}|}
\end{equation}

where \(\mathcal{M}^{\text{A}}, \mathcal{M}^{\text{B}} \subseteq \{1, \dots, D\}\) denote the sets of neurons (i.e., dimensions) selected by the thresholding step for a given functional network in each model. 

We compared LLaMA2-7B-Chat with its fine-tuned variants, as well as with training-data variants OpenLLaMA-7B and Amber (models that share the same architecture, but were trained on different datasets). We first report the highest spatial similarity (measured by IoU) between any pair of functional networks from the two models, along with the corresponding FNF value for that network pair. The results are shown in Table~\ref{tab::spatial}. As shown in Table~\ref{tab::spatial}, a higher spatial similarity (IoU) between functional networks generally corresponds to higher FNF values, especially for models that share a common origin. However, this pattern does not hold for non-homologous models. For example, in the case of Amber, the functional network pair with the highest IoU does not produce the highest FNF. Conversely, when we examine the network pair that achieves the highest FNF, their IoU is zero, indicating that there is no spatial overlap at all.

\begin{table}[h]
    \caption{The highest spatial similarity between LLaMA2-7B-chat and its fine-tuned variants and training-data variants.}
    \label{tab::spatial}
    \begin{center}
        \begin{tabular}{ccccccc}
            \toprule
            LLaMA2-7B & Vicuna & Chinesellama & Codellama & Llemma & OpenLLaMA & Amber\\ 
            \midrule
            IoU & 0.6115 & 0.6161 & 0.3971 & 0.3375 & 0.0735 & 0.1072\\ 
            FNF & 0.5215 & 0.9642 & 0.9184 & 0.8326 & 0.2061 & 0.2707\\
            \bottomrule
        \end{tabular}
    \end{center}
    \vskip -0.1in
\end{table}

To further investigate the relationship between spatial overlap and FNF, we directly applied LLaMA2-7B-chat’s 64 functional network masks to suspect models and computed their FNF to LLaMA2-7B-chat. We also applied each suspect model’s functional network masks to LLaMA2-7B-chat and measured the resulting FNF values. As shown in Table~\ref{tab::avg_fnf}, not all functional networks exhibit consistent activity, some inconsistent networks lower the average FNF. In particular, for models that are training-data variants, functional networks occupying the same spatial locations show completely unrelated activity patterns.

\begin{table}[h]
    \caption{The average FNF between LLaMA2-7B-chat and its fine-tuned variants and training-data variants. FNF-1 denotes the FNF computed by applying LLaMA2-7B-chat’s 64 functional network masks to the suspect model, while FNF-2 denotes the FNF obtained by applying the suspect model’s 64 functional network masks to LLaMA2-7B-chat. }
    \label{tab::avg_fnf}
    \begin{center}
        \begin{tabular}{ccccccc}
            \toprule
            LLaMA2-7B & Vicuna & Chinesellama & Codellama & Llemma & OpenLLaMA & Amber\\ 
            \midrule 
            FNF-1 & 0.3216 & 0.8875 & 0.6530 & 0.6005 & 0.0048 & -0.0010\\
            FNF-2 & 0.3090 & 0.8863 & 0.6561 & 0.5688 & 0.0039 & -0.0013\\
            \bottomrule
        \end{tabular}
    \end{center}
    \vskip -0.1in
\end{table}

\end{document}